\pdfoutput=1

\documentclass[11pt]{article}

\usepackage[final]{acl}
\usepackage{float}
\usepackage{times}
\usepackage{latexsym}

\usepackage[T1]{fontenc}

\usepackage[utf8]{inputenc}

\usepackage{microtype}

\usepackage{inconsolata}

\usepackage{csvsimple}

\usepackage{listings}
\usepackage{multirow}
\usepackage{booktabs}
\usepackage{graphicx}
\usepackage{subcaption}
\usepackage{tikz}
\usepackage{pgfplotstable}
\usepgfplotslibrary{colorbrewer}

\definecolor{codegreen}{rgb}{0,0.6,0}
\definecolor{codegray}{rgb}{0.5,0.5,0.5}
\definecolor{codepurple}{rgb}{0.58,0,0.82}
\definecolor{backcolour}{rgb}{0.95,0.95,0.92}
 
\lstdefinestyle{mystyle}{
    backgroundcolor=\color{backcolour},   
    commentstyle=\color{codegreen},
    keywordstyle=\color{magenta},
    numberstyle=\tiny\color{codegray},
    stringstyle=\color{codepurple},
    basicstyle=\footnotesize,
    breakatwhitespace=false,         
    breaklines=true,                 
    captionpos=b,                    
    keepspaces=true,                 
    numbers=none,                    
    numbersep=5pt,                  
    showspaces=false,                
    showstringspaces=false,
    showtabs=false,                  
    tabsize=2
}

\title{AGB-DE: A Corpus for the Automated Legal Assessment of Clauses in German Consumer Contracts}

\author{Daniel Braun\\
  University of Twente \\
  Department of High-tech Business\\
  and Entrepreneurship\\
  \texttt{d.braun@utwente.nl} \\\And
  Florian Matthes \\
  Technical University of Munich \\
  TUM School of Computation,\\
  Information and Technology\\  
  \texttt{matthes@tum.de} \\}

\begin{document}
\maketitle
\begin{abstract}
Legal tasks and datasets are often used as benchmarks for the capabilities of language models. However, openly available annotated datasets are rare. In this paper, we introduce AGB-DE, a corpus of 3,764 clauses from German consumer contracts that have been annotated and legally assessed by legal experts. Together with the data, we present a first baseline for the task of detecting potentially void clauses, comparing the performance of an SVM baseline with three fine-tuned open language models and the performance of GPT-3.5. Our results show the challenging nature of the task, with no approach exceeding an F1-score of 0.54. While the fine-tuned models often performed better with regard to precision, GPT-3.5 outperformed the other approaches with regard to recall. An analysis of the errors indicates that one of the main challenges could be the correct interpretation of complex clauses, rather than the decision boundaries of what is permissible and what is not.
\end{abstract}

\section{Introduction}

Standard form consumer contracts,  i.e. consumer contracts that are drafted unilaterally by a company, have huge significance for our economy, but also for consumer protection. Their review is a laborious task, performed by companies, law firms, governmental organizations, and non-governmental organizations (NGOs). In recent years, researchers have investigated different computational approaches to automate parts of the contract reviewing process (see Section \ref{sec:relwork}).

One of the big challenges for such research is the scarcity of contract data in general and annotated data in particular. Even more so for languages other than English. Annotating contracts with legal assessments is a laborious process that can only be performed by highly qualified experts and is therefore very expensive. Commercial providers of Large Language Models (LLMs), like OpenAI, but also the scientific community, increasingly use legal tasks to assess (or promote) the capabilities of LLMs. For many of the existing datasets and tasks, it is questionable whether LLMs like GPT-3.5 have not been trained on them, which would question the validity of evaluations performed on them \citep{balloccu2024leak}.

In this paper, we present a new corpus consisting of 3,764 clauses, 11,387 sentences, and 250,859 tokens from German consumer contracts. Each clause has been annotated by legal experts with a clause topic and whether the clause is valid or potentially void (see Section \ref{sec:annotation}). A contract clause is a section in a contract that is usually separated explicitly through formatting from other clauses and deals with a specific provision. The corpus contains a total of 8,582 labels and is available on GitHub\footnote{\url{https://github.com/DaBr01/AGB-DE}} and as Hugging Face dataset \footnote{\url{https://huggingface.co/datasets/d4br4/agb-de}}.

In addition, we present a baseline for the task of identifying potentially void clauses, comparing an SVM, three open language models in different sizes, and GPT-3.5. Our results show the challenging nature of the task, with no approach exceeding an F1-score of 0.54. The best-performing model, AGBert, is also available for download\footnote{\url{https://huggingface.co/d4br4/AGBert}}. While the fine-tuned models often performed better with regard to precision, GPT-3.5 outperformed the other approaches with regard to recall. An analysis of the errors indicates that one of the main challenges could be the correct interpretation of complex clauses, rather than the decision boundaries of what is permissible and what is not. The code that was used to prepare the datasets and train the models is also available on GitHub. We hope that the corpus will contribute to enabling future open and reproducible NLP research.

\section{Related Work}
\label{sec:relwork}
Legal tasks and datasets have become increasingly relevant for the evaluation of language models over the past years. From domain-specific models, like LEGAL-BERT \citep{chalkidis-etal-2020-legal}, to benchmark datasets, like LexGLUE \citep{chalkidis-etal-2022-lexglue}, culminating in the presentation of GPT-4 by OpenAI, where the alleged ``legal skills'' of the model have been one of the main communication points to show its improvement from the previous version \citep{achiam2023gpt}. Additionally, the question of whether a GPT model would be able to get through law school \citep{choi2021chatgpt} or pass the bar exam \citep{katz2023gpt} has been raised repeatedly in the last few years.

\subsection{Legal Datasets}
In general, there is a large number of legal data sets available. Mostly because many governments and institutions publish laws, court decisions, and similar legal documents digitally and under open licenses. However, only a small fraction of them has been manually annotated \citep{braun2023beg}. Many supervised NLP tasks in the legal domain therefore rely on corpora that inherently contain labels or where labels can be automatically derived. That is, for example, the case for translation where parallel corpora of the European Union can be used \citep{skadins-etal-2014-billions} or outcome prediction for court cases, where the final verdict can be easily extracted from the decision document \citep{chalkidis-etal-2019-neural}. Document types that are not regularly published by governments or institutions, like (consumer) contracts, are rare to find in corpora.

Manual annotation by legal experts is very expensive and therefore rare. While a number of such data sets exists (see Section \ref{sec:conscon} and \ref{sec:b2bcon}), particularly in English, we are not aware of any corpora in different languages that have a size comparable to the AGB-DE corpus, which consists of 3,764 legally annotated clauses from 93 contracts.

\subsubsection{Consumer Contracts}
\label{sec:conscon}
With 100 English Terms of Services and a total of 1,715 annotated clauses, the corpus provided by \citet{ruggeri2022detecting} is one of the biggest of its kind. Clauses are classified into fair and unfair clauses and among the unfair clauses five major categories are distinguished. \citet{drawzeski-etal-2021-corpus} presented a similar sized corpus of 100 Terms of Services, however, consisting of only 25 distinct contracts each of which is available in four languages (English, Italian, German, and Polish). 

Another large corpus of consumer contracts is the OP-115 Corpus by \citet{wilson-etal-2016-creation} which consists of 115 privacy policies from websites that have been annotated by law students with data practices that occur in the text. The MAPP corpus is an even larger privacy policy corpus with 155 privacy policies from mobile applications \citep{arora-etal-2022-tale}, which have been annotated in a similar fashion. Both contain only English contracts.

\subsubsection{B2B Contracts}
\label{sec:b2bcon}
In the space of commercial (business-to-business) contracts, larger datasets are available. The CUAD dataset by \citet{hendrycks2021cuad} consists of 510 English commercial contracts in which 41 different types of legal clauses have been annotated. \citet{10.1145/3086512.3086515} provide a data set of 993 contracts that have been labeled with clause headings and 2,461 contracts that have been annotated with other types of contract elements by law students. Unlike, for example, the data set by \citet{ruggeri2022detecting}, but also the AGB-DE corpus, these datasets were not annotated with a legal judgment but rather with regard to the topic of individual clauses.

\subsection{Legal Assessment of Contract Clauses}
Some of the above-mentioned data sets, but also other, mostly non-published data sets, have been used to train different NLP models in order to predict whether a given contract clause is valid or not. This is also the task we mainly had in mind when we built the AGB-DE corpus. \citet{ruggeri2022detecting} used a Memory-Augmented Neural Network to detect unfair clauses in Terms of Services and reported an accuracy of 0.526. \citet{braun-matthes-2021-nlp} used a fine-tuned BERT model to predict void clauses in Terms and Conditions of online shops and reported an accuracy of 0.9. \citet{9218152} used an SVM to detect missing clauses in privacy policies and reported a precision of 0.85 and a recall of 0.96. Most recently, \citet{martin2024better} presented a study that compares the performance of, among others, GPT4-32k with junior lawyers in locating legal issues in contracts. They report that GPT4 is not only faster and cheaper but also better (F1-score of 0.74) than junior lawyers (F1-score of 0.667).

\section{Corpus}
The corpus consists of 3,764 clauses from 93 standard form consumer contracts. In this section, the construction of the corpus is described. A detailed datasheet \citep{gebru2021datasheets} for the corpus can be found in Appendix \ref{sec:datasheet}.

\subsection{Data}
The data was collected in 2021 and 2022 and annotated between 2021 and 2023. It consists of German\footnote{The contracts are not only written in the German language but also tailored to the German market and its regulations.} standard form consumer contracts that are available online, such as Terms and Conditions of online shops, fitness studios, and telecommunication providers. The data was collected by the annotators, consumer protection lawyers (see Section \ref{sec:annotation} for more details), based on their professional interests. Each clause from each contract was manually copied into an Excel file together with the title of the clause and the URL to the contract text. In total, 93 contracts have been collected in that way.

\subsection{Annotation}
\label{sec:annotation}

The dataset was annotated by five fully-qualified lawyers from two German NGOs with a focus on consumer protection. Each of the annotators had multiple years of experience in consumer protection law and consumer counseling. The annotations were made in an Excel file, which annotators preferred over dedicated annotation tools.

\subsubsection{Topics}
First, one or multiple topic labels were added to each clause. Subsequently, subtopic labels could be added to further specify the content of a clause. For this classification, we used the taxonomy introduced by \citet{braun-matthes-2022-clause}, which consists of 23 topic labels and 37 subtopic labels (see Appendix \ref{sec:apptax} for a list of the available labels). While each clause had to be annotated with at least one topic label, the annotators were instructed to only add subtopics where they found it fitting.

\subsubsection{Validity}
Afterwards, each clause was legally assessed by the expert annotators. A clause in a contract is \textit{void}, i.e. cannot be enforced by the parties of the contract, if it contradicts governing law. Whether a clause is actually void depends on many things, including, in some cases, whether one of the parties is a consumer or whether both parties are businesses. The final decision on whether a specific clause in specific circumstances is actually void can only be made by a court of law. Therefore, the instruction for the annotators was to label a clause as \textit{potentially} void, if they think a consumer residing in Germany could successfully challenge the clause in court. For the remainder of this paper, if we say a clause is void, we mean that it was annotated as potentially void by the expert annotators. In addition to the assessment itself, which is binary (1 - potentially void, 0 - valid), the annotators can add a comment in the Excel file explaining their decision. By the wish of the annotators, these explanations are not part of the published corpus. It is also worth re-highlighting that the annotators work for NGOs that are dedicated to advocate for consumer rights and their interpretation of the law might therefore be more consumer-friendly than a lawyer who works for a big corporation.

Initially, a small subset of the data was annotated by two experts (one from each organization). In this initial annotation, the legal annotations were in agreement in 76\% of the cases. Based on these initial annotations, the experts discussed and aligned their annotation strategies to increase the consistency of the annotations. During this process, two patterns underlying the disagreement became apparent. All disagreements that were found in this phase were based on a disagreement about the interpretation of laws or court rulings, rather than a disagreement about the interpretation of a clause, i.e. the text of the contract. The second pattern was that many of the disagreements were based on laws that use vague legal terms. Laws are often formulated vaguely on purpose. The German Civil Code for example deems clauses void that provide ``unreasonably long payment deadlines'' (§308 No. 1a). What might seem like bad law-making is done to make laws ``future proof''. Whether a payment deadline is unreasonably long, for example, changed significantly between the 1970s, when letters and bank transfers still took multiple days and today. It is up to courts to interpret these terms and these interpretations can also change over time. To increase the alignment of annotators, the annotations guidelines were extended with agreed interpretations of relevant legal provisions that contain such vague legal terms. Additionally, as a ``catch-all'' solution, a rule was introduced that when in doubt, e.g. because different courts ruled differently, the annotators will always use the more consumer-friendly interpretation. For the subsequent annotation of the complete corpus, each instance was annotated by one annotator. While it would have been preferable to have multiple annotators per instance, that was not feasible due to cost reasons. The total costs of the data collection and annotation for the AGB-DE corpus were approximately 110.000 EUR.

\subsection{Anonymization}
For ethical and legal reasons, we decided to anonymize the dataset before publishing it. While the original data does not contain any information from individuals, it does contain (publicly accessible) information from companies, like phone numbers, addresses, and tax IDs. We took multiple steps to remove this data from the contracts in order to make it harder to identify the company that drafted the contract. Companies can and do change their contracts over time, so we want to avoid consumers finding and reading an outdated version of a contract. Additionally, it is not unlikely that one or multiple assessments made by the annotators would not hold up in a court of law, either due to an unconscious mistake or due to the aforementioned bias with regard to the interpretation of the law. Wrongfully claiming a company uses void terms can potentially be harmful to their business and could implicate liabilities. The other way around, wrongfully claiming a clause is valid could potentially harm consumers if they rely on that assessment. Therefore we implemented ten anonymization steps:
\begin{enumerate}
    \item Remove all clauses with the topic label \texttt{party} from the corpus (these clauses consist only of information about the contracting party, i.e. the company)
    \item Replace all email addresses with ``hello@example.com'' using regular expressions (regex)
    \item Replace all URLs with ``www.example.com'' using regex
    \item Replace all international bank account numbers (IBANs) with ``DE75512108001245126199'' using regex
    \item Replace all tax IDs with DE398517849 using regex
    \item Replace all phone numbers with 00 00 12345678 using regex
    \item Replace all ZIP codes with 00000 using regex
    \item Replace all names of companies and organizations with ``<<NAME>>'' using Named Entity Recognition (NER)
    \item Replace all city names with ``<<STADT>>'' (German for city) using NER
    \item Replace all street names with ``<<STRASSE>>'' (German for street) using NER
\end{enumerate}

While the first seven steps turned out to work very well and straight-forward (in total 84 \texttt{party} clauses have been removed and 120 email addresses, 231 URLs, 2 IBANs, 117 phone numbers, and 279 ZIP codes have been replaced), many of the available standard NER libraries turned out to not work very well for the texts. In the end, the FLAIR library \citep{akbik-etal-2019-flair} with the \texttt{ner-german-legal} model \citep{leitner2019fine} turned out to be most suitable. With the help of the model, we were able to replace 724 names of organizations and companies, 418 city names, and 53 street names. However, a manual inspection revealed that an additional, manual, anonymization round was necessary. In this manual process an additional 1,338 company names, 38 city names, and 85 streets have been removed.

In order to avoid over-anonymization which could potentially result in decreased classification performance, a list of organizations and URLs were explicitly excluded from being removed or replaced. The list mainly included political bodies like the European Union and their URLs, shipping companies and their URLs, and payment provider and their URLs. An excerpt from the final corpus is shown in Table \ref{tab:example}.

\begin{table*}
    \centering
    \begin{tabular}{|r r l p{2cm} p{3.8cm} l p{1.8cm} r|}
    \hline
         \textbf{Id} & \textbf{Con.} &\textbf{Lang} & \textbf{Title} & \textbf{Text} & \textbf{Topics} & \textbf{Subtopics} & \textbf{Void}  \\\hline
         10 & 1 & de & 2. Widerrufsbelehrung & Sie tragen keine Kosten für die Rücksendung der Ware. & withdrawal & withdrawal: shippingCosts & 0\\\hline
         124 & 3 & de & 11. Geltungsbedingungen der AGB & Anstelle der unwirksamen Vorschrift gilt eine Regelung, die der mit der unwirksamen Vorschrift verfolgten wirtschaftlichen Zwecksetzung am nächsten kommt.          
 & severability & & 1 \\\hline
 127 & 4 & de & 1. Allgemeines &1.4. Mit der Bestellung auf dieser Website bestätigen Sie, dass Sie volljährig und rechtsfähig sind, einen Verbrauchervertrag abzuschließen." & age & & 0\\\hline
 215 & 9 & de & §6 - EIGENTUMSVORBEHALT & Die von <<NAME>> gelieferte Ware verbleibt bis zur vollständigen Bezahlung Eigentum von <<NAME>> & ret.OfTitle & &0\\\hline
    \end{tabular}
    \caption{Excerpt from the corpus}
    \label{tab:example}
\end{table*}

\section{Corpus Analysis}

The corpus consists of 93 contracts with 3,764 clauses (an average of 40 clauses per contract), which contain 11,387 sentences (avg. of 3 sentences per clause) and 250,859 tokens (avg. of 22 per sentence). Out of the 3,764 clauses present in the corpus, 179 (or 4.8\%) have been annotated as potentially void. That is comparable, although slightly lower, than the 6\% reported by \citet{braun-matthes-2021-nlp} on a much smaller dataset of 24 contracts. While that results in a corpus that is imbalanced, we believe it to be a realistic reflection of reality, where void clauses are also significantly less frequent than void clauses.

Table \ref{tab:topicdist} shows how the clauses are distributed among the topics and the percentage of potentially void clauses in each topic. Since a clause can belong to multiple topics and subtopics, the sum of the labels is greater than the number of clauses. 30 clauses have been annotated with more than one topic. Out of those 30, one clause has been annotated with three topics and one clause has been annotated with four topic labels, the other 28 have been annotated with two topic labels. Out of the 3,764 clauses, only 1,078 have been annotated with a subtopic. Partially, we believe that to be the result of the fact that it was not mandatory for the annotators to add a subtopic and that the focus of the annotation was clearly on the legal assessment. In total, the corpus contains 8,582 labels.

An analysis of the distribution of void clauses in relation to the topic shows that there are some classes which are particularly prone to be seen as potentially void by the experts. A deeper analysis, together with the experts, revealed that these are very often types of clauses that are particularly strictly regulated. The class \texttt{changes}, for example, captures clauses that relate to changes that are made to the contract after it came into force. Given that standard form contracts are already considered to ``reflect an imbalance of contracting power'' \citep{braun2019potential}, it is not surprising that the possibilities for a company to change them after the customer agreed are very strictly regulated. Therefore, it makes sense that such clauses are exceedingly considered void by the annotators. Similar reasoning can be applied to other topics, like liability or severability clauses.

\begin{table}
\small
\begin{tabular}{|l r r |}
\hline\textbf{Label} & \textbf{Amount} & \textbf{Void (\%)}
\\\hline
\begin{filecontents*}{files/corp_analysis.csv}
label,clauses_total,share_void
age,20,0.00
applicability,148,2.03
applicableLaw,87,3.45
arbitration,97,1.03
changes,9,11.11
codeOfConduct,29,0.00
conclusionOfContract (cOc),557,5.92
cOc:binding,50,0.00
cOc:changeOfOrder,1,0.00
cOc:definition,39,0.00
cOc:restrictions,13,0.00
cOc:steps,59,0.00
cOc:withdrawal,17,0.00
contractLanguage,41,0.00
delivery,475,7.16
delivery:brokenPackaging,19,0.00
delivery:costs,53,0.00
delivery:customs,2,0.00
delivery:destination,18,0.00
delivery:methods,26,0.00
delivery:partial,22,0.00
delivery:time,1,0.00
description,46,0.00
disposal,36,0.00
intellectualProperty,39,0.00
language,9,11.11
liability,211,9.00
party,0,0.00
payment,642,6.07
payment:fee,6,0.00
payment:late,27,0.00
payment:loyalty,1,0.00
payment:methods,289,0.00
payment:restraint,5,0.00
payment:vouchers,124,0.00
personalData,115,0.87
personalData:cookies,6,0.00
personalData:duration,0,0.00
personalData:information,1,0.00
personalData:reason,1,0.00
personalData:update,0,0.00
personalData:usage,1,0.00
placeOfJurisdiction,53,1.89
prices,147,1.36
prices:currency,6,0.00
prices:vat,36,0.00
retentionOfTitle,125,2.40
severability,35,11.43
textStorage,57,1.75
warranty,314,6.37
warranty:options,4,0.00
warranty:period,10,0.00
withdrawal,506,3.75
withdrawal:compensation,7,0.00
withdrawal:effects,45,0.00
withdrawal:exclusion,51,0.00
withdrawal:form,25,0.00
withdrawal:model,4,0.00
withdrawal:period,33,0.00
withdrawal:shippingCosts,11,0.00
withdrawal:shippingMethod,7,0.00
Total lvl 1,3798,4.80
Total lvl 2,1020,
\end{filecontents*}
\csvreader[before line=\\,before first line=]{files/corp_analysis.csv}{}{\csvcoli & \csvcolii & \csvcoliii}%
\\\hline
\end{tabular}
\caption{Distribution of topics and void clauses}
\label{tab:topicdist}
\end{table}

\section{Automated Legal Assessment}

In order to present a baseline for the main task for which the corpus was designed and evaluate the difficulty of the task, we compared different language models and an SVM for the classification of clauses into (potentially) void and valid.

\subsection{Data Split}
To do so, we created a dataset from the corpus, which is split into a training set of 80\% (3,004 clauses) and a test set of 20\% (755 clauses)\footnote{\url{https://huggingface.co/datasets/d4br4/agb-de}}. We stratified the data split by both the topic labels and the legal assessment to guarantee an equal representation of each class in both datasets. Five clauses from the original corpus were removed in this dataset because they were the only void instances of their clause type in the corpus, and it was, therefore, not possible to split them in the above-described fashion.

\subsection{Undersampling}
Because it was clear that the dataset would be challenging because of its imbalanced nature, we created a second dataset\footnote{\url{https://huggingface.co/datasets/d4br4/agb-de-under}}, in which we used undersampling \citep{liu2008exploratory} to remove data from classes that are over-represented. In particular, the number of instances from each combination of topic and validity was limited to 100. I.e., if there were 40 void clauses of one topic and 120 valid clauses of the same topic, only 100 of the valid clauses would go into this second dataset. In this way, we ended up with a dataset that consists of 1,362 (80\%) clauses for training and 345 (20\%) for testing. The split of the data remained the same, i.e. we only removed data but did not change the distribution of existing data between the training and test sets. For easier distinction, we will refer to the first larger dataset as \texttt{agb-de} and to the undersampled dataset as \texttt{agb-de-under}.

\subsection{Models}
We used both datasets to train an SVM that uses tf-idf vectors of the clauses as input and fine-tune and evaluate four different language models:
\begin{itemize}
    \item The BERT model \texttt{bert-base-german-cased} \citep{chan-etal-2020-germans}, which was also trained on the Open Legal Data dataset that consist of more than 100,000 German legal documents \citep{10.1145/3383583.3398616}
    \item The multilingual RoBERTa model \texttt{xlm-roberta-base} \citep{DBLP:journals/corr/abs-1911-02116}
    \item The German GPT2 model \texttt{gerpt2} \citep{Minixhofer_GerPT2_German_large_2020}
    \item And finally \texttt{gpt-3.5-turbo-0125}, which was evaluated in a ``zero-shot'' fashion without fine-tuning
\end{itemize}

For the fine-tuning of the three open models, we conducted a manual hyperparameter search, starting with the standard hyperparameters for each model. The final hyperparameters that were used for the fine-tuning can be found in Appendix \ref{sec:appfine}, as well as in the training code that is published alongside the corpus. In order to further address the imbalance of the dataset, we also used a tailored loss function. For the fine-tuning of all models, we used class weights of 1.0 for valid and 100.0 for void clauses in the loss function. The prompt and API call used for the evaluation of GPT-3.5 can be found in Appendix \ref{sec:appprompt}.

\section{Evaluation}
\begin{table*}[]
    \centering
    \begin{tabular}{|l|l|r|r|r|}
    \hline
        \textbf{Dataset} &  \textbf{Model} & \textbf{Precision} & \textbf{Recall} & \textbf{F1-score}\\\hline
        \multirow{5}{*}{\texttt{agb-de}} & \texttt{svm} & 0.37 & 0.27 & 0.31\\ 
        & \texttt{bert-base-german-cased} & 0.50 & 0.27 & \textbf{0.35}\\
        & \texttt{xlm-roberta-base} & 0.00 & 0.00 & 0.00\\
        & \texttt{gerpt2} & \textbf{0.71} & 0.14 & 0.23\\
        & \texttt{gpt-3.5-turbo-0125} & 0.06 & \textbf{0.92} & 0.11\\\hline
        \multirow{4}{*}{\texttt{agb-de-under}} & \texttt{svm} & 0.40 & 0.32 & 0.36 \\
        & \texttt{bert-base-german-cased} & 0.51 & 0.57 & \textbf{0.54}\\
        & \texttt{xlm-roberta-base} & \textbf{0.75} & 0.08 & 0.15\\
        & \texttt{gerpt2} & 0.64 & 0.43 & 0.52\\
        & \texttt{gpt-3.5-turbo-0125} & 0.13 & \textbf{0.92} & 0.22\\\hline
    \end{tabular}
    \caption{Results of the evaluation, best performance on a dataset is highlighted in bold}
    \label{tab:eval}
\end{table*}

The results of the evaluation are shown in Table \ref{tab:eval}. The main metric that we focused on for the evaluation is the F1-score, which provides a balance between the two unequally distributed classes.

For the \texttt{agb-de} dataset, none of the models was able to handle the very imbalanced dataset well. The best-performing model on the dataset was the fine-tuned BERT model. While RoBERTa is usually better than BERT in handling imbalanced data \citep{younes-mathiak-2022-handling}, the RoBERTa model that we fine-tuned is a multilingual model, while the BERT model that we fine-tuned was specifically pre-trained on German legal data. The second best performance was achieved by the SVM with an F1-score of 0.31. For GPT-3.5, there was not much difference in the performance independent of the dataset: For both datasets, GPT-3.5 achieved the highest recall, at the cost of a very low precision. I.e., GPT-3.5 falsely classified the majority of clauses as potentially void (see also Figure \ref{fig:gptconf}).

All models performed better on the \texttt{agb-de-under} dataset, showing that undersampling is a suitable strategy for the corpus. BERT again performed best with an F1-score of 0.54. To test whether the improvement performance on the undersampled dataset was caused by a better model and not just by the fact that, by chance, difficult items got removed from the test set, we also tested all models trained on the undersampled dataset on the original test set from the \texttt{agb-de} dataset. The results of this evaluation are shown in Appendix \ref{sec:evalout}. Except for the SVM, all models performed better when being trained on the undersampled dataset and evaluated on the full dataset compared to being trained and tested on the full dataset, indicating that removing excessive imbalanced training data can indeed lead to better models on this dataset.

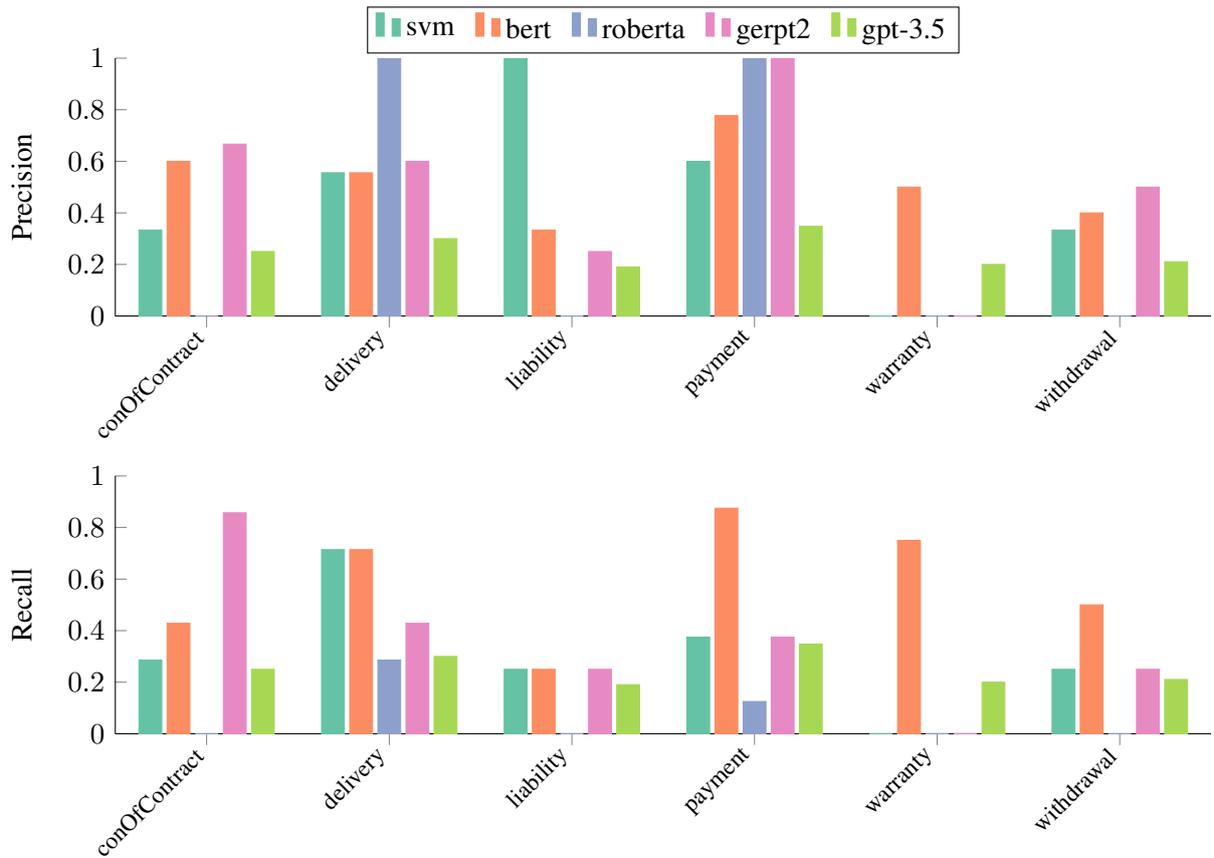
\begin{figure*}
    \centering
\pgfplotstableread[col sep=comma,]{files/error_svm.csv}\svm
\pgfplotstableread[col sep=comma,]{files/error_bert.csv}\bert
\pgfplotstableread[col sep=comma,]{files/error_roberta.csv}\roberta
\pgfplotstableread[col sep=comma,]{files/error_gerpt.csv}\gerpt
\pgfplotstableread[col sep=comma,]{files/error_gpt3.csv}\gpt
\begin{subfigure}{\textwidth}
\begin{tikzpicture}
\begin{axis}[
    ybar,
    bar width=0.3cm,
    width=\textwidth,
    height=5cm,
    ymin = 0,
    ymax = 1,
    xtick=data,
    xticklabels from table={\svm}{Topic},
    x tick label style={font=\small, rotate=45, anchor=east},
    ylabel={Precision},     
    ybar legend,
        cycle list/Set2,
every axis plot/.append style={fill},
     axis x line*=bottom,
     axis y line*=left,
    legend style={at={(0.5,1.2)},
    legend style={/tikz/every even column/.append style={column sep=0.2cm}},
 anchor=north,legend columns=-1},
    ]
    \addplot table [x expr=\coordindex, y={Precision}]{\svm};
    \addplot table [x expr=\coordindex, y={Precision}]{\bert};
    \addplot table [x expr=\coordindex, y={Precision}]{\roberta};
    \addplot table [x expr=\coordindex, y={Precision}]{\gerpt};
    \addplot table [x expr=\coordindex, y={Precision}]{\gpt};
    \legend{svm, bert, roberta, gerpt2, gpt-3.5}
\end{axis}
\end{tikzpicture}
\label{fig:error}
\end{subfigure}
\begin{subfigure}{\textwidth}
\begin{tikzpicture}
\begin{axis}[
    ybar,
    bar width=.3cm,
    width=\textwidth,
    height=5cm,
    ymin = 0,
    ymax = 1,
    xtick=data,
     axis x line*=bottom,
     axis y line*=left,
    xticklabels from table={\svm}{Topic},
    x tick label style={font=\small, rotate=45, anchor=east},
    ylabel={Recall},     
    cycle list/Set2,
every axis plot/.append style={fill}
    ]
    \addplot table [x expr=\coordindex, y={Recall}]{\svm};
    \addplot table [x expr=\coordindex, y={Recall}]{\bert};
    \addplot table [x expr=\coordindex, y={Recall}]{\roberta};
    \addplot table [x expr=\coordindex, y={Recall}]{\gerpt};
    \addplot table [x expr=\coordindex, y={Precision}]{\gpt};
\end{axis}
\end{tikzpicture}
\end{subfigure}

\caption{Precision and recall per topic on the \texttt{agb-de-under} dataset\label{fig:error}}
\end{figure*}

\section{Error Analysis}

\begin{figure*}[ht]
\begin{subfigure}{0.25\textwidth}
    \begin{tabular}{@{}cc|cc@{}}
    \multicolumn{1}{c}{} &\multicolumn{1}{c}{} &\multicolumn{2}{c}{Predicted} \\ 
    \multicolumn{1}{c}{} & 
    \multicolumn{1}{c|}{} & 
    \multicolumn{1}{c}{valid} & 
    \multicolumn{1}{c}{void} \\ 
    \cline{2-4}
    \multirow[c]{2}{*}{\rotatebox[origin=tr]{90}{Actual}}
    & valid  & 288 & 20   \\[1.5ex]
    & void  & 16 & 21\\ 
    \cline{2-4}
    \end{tabular}
    \caption{\texttt{bert-base-german-cased}}
\end{subfigure}
\begin{subfigure}{0.24\textwidth}
    \begin{tabular}{@{}cc|cc@{}}
    \multicolumn{1}{c}{} &\multicolumn{1}{c}{} &\multicolumn{2}{c}{Predicted} \\ 
    \multicolumn{1}{c}{} & 
    \multicolumn{1}{c|}{} & 
    \multicolumn{1}{c}{valid} & 
    \multicolumn{1}{c}{void} \\ 
    \cline{2-4}
    \multirow[c]{2}{*}{}
    & valid  & 307 & 1   \\[1.5ex]
    & void  & 34 & 3\\ 
    \cline{2-4}
    \end{tabular}
    \caption{\texttt{xlm-roberta-base}}
\end{subfigure}
\begin{subfigure}{0.24\textwidth}
    \begin{tabular}{@{}cc|cc@{}}
    \multicolumn{1}{c}{} &\multicolumn{1}{c}{} &\multicolumn{2}{c}{Predicted} \\ 
    \multicolumn{1}{c}{} & 
    \multicolumn{1}{c|}{} & 
    \multicolumn{1}{c}{valid} & 
    \multicolumn{1}{c}{void} \\ 
    \cline{2-4}
    \multirow[c]{2}{*}{}
    & valid  & 299 & 9   \\[1.5ex]
    & void  & 21 & 16\\ 
    \cline{2-4}
    \end{tabular}
    \caption{\texttt{gerpt2}}
\end{subfigure}
\begin{subfigure}{0.24\textwidth}
    \begin{tabular}{@{}cc|cc@{}}
    \multicolumn{1}{c}{} &\multicolumn{1}{c}{} &\multicolumn{2}{c}{Predicted} \\ 
    \multicolumn{1}{c}{} & 
    \multicolumn{1}{c|}{} & 
    \multicolumn{1}{c}{valid} & 
    \multicolumn{1}{c}{void} \\ 
    \cline{2-4}
    \multirow[c]{2}{*}{}
    & valid  & 71 & 237   \\[1.5ex]
    & void  & 3   & 34 \\ 
    \cline{2-4}
    \end{tabular}
    \caption{\texttt{gpt-3.5-turbo-0125}\label{fig:gptconf}}
\end{subfigure}
\caption{Confusion matrices for the evaluation on the \texttt{agb-de-under} dataset}
\end{figure*}

Figure \ref{fig:error} shows precision and recall for each model on the undersampled dataset for the largest topics in the corpus. The numbers show that the models perform very differently for the individual topics.

For liability, for example, no model was able to achieve a recall higher than 0.25. With 11\% of void clauses, liability clauses are most frequently void among the large topics. For the expert annotators, liability clauses were relatively easy to annotate, because of very explicit regulations. § 309 No. 7 of the German Civil Code, for example, explicitly states that ``an exclusion or limitation of liability for damage from injury to life, limb or health due to negligent breach of duty by the user'' is void. Linguistically, however, an analysis of the liability clauses shows that they are often complicated, with multiple explicit inclusions and exclusions in one sentence, which could be a reason for the poor performance of most models.

While clauses of a certain topic can be void for different reasons, for some topics there are predominant patterns. Warranty clauses, in the corpus, for example, are predominantly void because they restrict the warranty in cases where defects are not reported within a specified (too short) time frame. Similarly, payment clauses are predominantly void because they introduce excessive fees for late payments. The BERT model was better at picking up these more repetitive patterns compared to the more nuanced other topics. The GPT-2 model achieved the highest precision for payment clauses, which have the largest share in the corpus, and therefore heavily influence the overall result. Overall, the performance differences between classes could indicate that using the topic labels as an additional feature could improve classification performance. 

The performance of GPT-3.5 differs from the other models as it was more prone to generating false positives, i.e. flagging clauses as void that are actually valid. The prompt we used for GPT-3.5 not only asked it to perform the classification into potentially valid and void but also asked the model to provide an ``explanation''\footnote{The ``explanations'' can be found in the GitHub repository.}. While the texts provided are not an explanation in the sense that they make the reasons for the assessment transparent, it is still interesting to see that the issues described are most of the time correct, however, the assessment is still wrong, in both the text and the annotation. For a clause about withdrawals (corpus ID 5), GPT-3.5 for example ``explains'': \textit{``The clause is potentially invalid as it could unfairly disadvantage the consumer. According to the law, the consumer must be able to exercise his right of withdrawal clearly and conspicuously, without any additional hurdles or conditions being imposed. By specifying here that registering a return under My Account is considered a revocation, this could limit the consumer's ability to exercise their right of revocation. It is important that the consumer can exercise his right of withdrawal without additional obligations.''} (see Appendix \ref{sec:apex1} for the original German text) While the text by itself would be factually correct, it is not applicable to the clause in question. The clause clearly describes other ways to exercise the right to revoke a contract and explicitly states that it is not mandatory to use the ``My Account'' feature. In line with the text, the clause was falsely labeled as potentially void by GPT-3.5.

Similarly, for the clause ``If the customer is a merchant [...] the place of jurisdiction for all disputes arising from this contract is the court responsible for the seller's place of business [...].'' (corpus ID 3212) GPT-3.5 concludes the clause is potentially void, because \textit{``The clause puts the consumer at an unreasonable disadvantage because this clause binds the consumer to the company's place of jurisdiction'}'. The legal reasoning of that is again correct, however, not applicable to the clause, because the clause only applies to merchants in which case it is valid. To test whether GPT-3.5 is aware that the clause is valid for merchants, we adapted the prompt to say that the system should imagine being a lawyer for a merchant (instead of a consumer) and indeed the response was ``The clause is unlikely to be potentially invalid if the customer is an entrepreneur.'' More often than not, the main challenge for GPT-3.5 seems not to be missing legal ``knowledge'', but the incorrect interpretation of the clause, especially in cases where the clause contains elements that are optional or not applicable to all customers.

Another typical problem of LLMs in general can be seen in the explanation generated for the clause with the corpus ID 10. Here, the model generates the following text: \textit{``According to Section 357 of the German Civil Code (BGB), a consumer may not be charged the costs of return shipping in the event of a revocation. Therefore, a clause requiring the consumer to bear the costs of return shipping is potentially invalid.''} (see Appendix \ref{sec:apex2}). Until 2014, it was indeed the case that \textit{``a consumer may not be charged the costs of return shipping in the event of a revocation''} (at least for purchases over 40 EUR). However, today, §357 BGB explicitly states the opposite, i.e. that customers have to bear the costs of return shipping. However, texts that relate to the old legislation are probably more frequent in the data GPT-3.5 was trained on, which is data up to 2021.

\section{Conclusion}

In this paper, we have introduced the AGB-DE corpus, consisting of 3,764 clauses from German consumer contracts that have been annotated by legal experts. In addition to the corpus, we presented two datasets that have been derived from the corpus and have been split into training and test sets. The datasets can easily be used to train or fine-tune machine learning models. We have used these datasets to provide a benchmark on the corpus for the task of classifying whether a clause is valid or potentially void.

For the dataset that is representative of the distribution in the corpus, we showed that language models struggle with the imbalanced data and could not achieve an F1-score above 0.35. For the second dataset, which is more balanced through undersampling, we showed that open models like BERT and GPT2 were able to better identify void clauses achieving an F-1 score of 0.54 and 0.52. On both datasets, open models outperformed GPT-3.5 with regard to F1-score, which generated a huge amount of false positives leading to an F1-score of 0.11 on the more imbalanced and 0.22 on the less imbalanced dataset.

\section*{Ethics}
False legal statements always have to potential to cause harm. While we worked with experienced experts who carefully decided on each annotation, it is always possible that the dataset contains errors. Claiming falsely that a clause is void could potentially have negative impacts on a business, claiming falsely that a clause is valid could potentially have a negative impact on consumers. One measurement we took to avoid such impacts is anonymising the clauses, in order to make it harder to connect them with a specific business. Many clauses, like severability or liability clauses, are highly standardised and often directly drawn from boilerplate contracts. In such cases removing the explicit identifiers is sufficient to make them completely anonymous. In other cases, particularly for example in clauses about bonus and rewards programs, even after removing the explicit identifiers, clauses can be so specific that they can be traced back to individual businesses, if they still use the same clause. We believe that the practical impact of that is rather limited, because retrieving the information about a specific clause from the dataset is not something that a significant number of potential customers is likely going to do. We added an explanation text to the readme of the published version of the corpus to highlight that the annotations are based on the perspective of individual experts and not based on a concrete court ruling and therefore not legally binding. Overall we believe that the publication of the dataset can help to address an existing imbalance of power between customers and companies and thereby, together with the anonymisation strategy, warrants the small but existing risk.

At the same time, being aware that a company uses potentially void clauses that disadvantage consumers and not doing anything about it could be seen as unethical. Therefore, the NGOs we worked with did take legal steps if they encountered clauses during the annotation that they deemed so disadvantageous for consumers that legal steps were necessary.

In general, the dataset consists of publicly accessible data that is usually carefully drafted by lawyers. We did not encounter any instances within the dataset that we deem problematic from an ethical perspective.

\section*{Limitations}
\label{sec:limitations}

Due to practical restrictions, the presented dataset has several limitations:

\begin{itemize}
    \item The dataset was annotated from a consumer protection perspective and therefore is most likely biased towards interpreting existing regulations in a consumer-friendly way.
    \item The instances in the dataset were only annotated by one expert. Legal decision-making involves uncertainty and interpretation, therefore it would have been desirable to have each instance annotated by multiple experts.
    \item The assessments in the corpus are based on the legal regulations at the time of the annotation (2021-2023), models trained on newer data might correctly make different predictions based on legislative changes or new court decisions.
    \item While we believe the dataset to be a somewhat representative mapping of the real world, that also means that is imbalanced and contains a relatively small amount of void clauses.
\end{itemize}

Due to practical restrictions, the presented evaluation has several limitations:
\begin{itemize}
    \item Using cloud-based models like GPT-3.5 always poses a threat to the reproducibility of the results, through using an explicitly versioned instance of the model we try to minimize the risk.
    \item Arguably comparing models that were fine-tuned on a specific task with a zero-shot prompt approach is an unequal comparison. Future strategies to improve the GPT-3.5 performance could include using few-shot approaches or also prompt engineering. While, given the error results, the few-shot approach might have limited effect, we believe that prompt engineering might be effective in reducing the large number of false positives.
\end{itemize}

\section*{Acknowledgments}
The data collection and annotation was supported by funds of the Federal Ministry of Justice and Consumer Protection (BMJV) based on a decision of the Parliament of the Federal Republic of Germany via the Federal Office for Agriculture and Food (BLE) under the innovation support
programme.

\bibliography{anthology,custom}

\begin{thebibliography}{29}
\expandafter\ifx\csname natexlab\endcsname\relax\def\natexlab#1{#1}\fi

\bibitem[{Achiam et~al.(2023)Achiam, Adler, Agarwal, Ahmad, Akkaya, Aleman, Almeida, Altenschmidt, Altman, Anadkat et~al.}]{achiam2023gpt}
Josh Achiam, Steven Adler, Sandhini Agarwal, Lama Ahmad, Ilge Akkaya, Florencia~Leoni Aleman, Diogo Almeida, Janko Altenschmidt, Sam Altman, Shyamal Anadkat, et~al. 2023.
\newblock Gpt-4 technical report.
\newblock \emph{arXiv preprint arXiv:2303.08774}.

\bibitem[{Akbik et~al.(2019)Akbik, Bergmann, Blythe, Rasul, Schweter, and Vollgraf}]{akbik-etal-2019-flair}
Alan Akbik, Tanja Bergmann, Duncan Blythe, Kashif Rasul, Stefan Schweter, and Roland Vollgraf. 2019.
\newblock \href {https://doi.org/10.18653/v1/N19-4010} {{FLAIR}: An easy-to-use framework for state-of-the-art {NLP}}.
\newblock In \emph{Proceedings of the 2019 Conference of the North {A}merican Chapter of the Association for Computational Linguistics (Demonstrations)}, pages 54--59, Minneapolis, Minnesota. Association for Computational Linguistics.

\bibitem[{Arora et~al.(2022)Arora, Hosseini, Utz, Bannihatti~Kumar, Dhellemmes, Ravichander, Story, Mangat, Chen, Degeling, Norton, Hupperich, Wilson, and Sadeh}]{arora-etal-2022-tale}
Siddhant Arora, Henry Hosseini, Christine Utz, Vinayshekhar Bannihatti~Kumar, Tristan Dhellemmes, Abhilasha Ravichander, Peter Story, Jasmine Mangat, Rex Chen, Martin Degeling, Thomas Norton, Thomas Hupperich, Shomir Wilson, and Norman Sadeh. 2022.
\newblock \href {https://aclanthology.org/2022.lrec-1.585} {A tale of two regulatory regimes: Creation and analysis of a bilingual privacy policy corpus}.
\newblock In \emph{Proceedings of the Thirteenth Language Resources and Evaluation Conference}, pages 5460--5472, Marseille, France. European Language Resources Association.

\bibitem[{Balloccu et~al.(2024)Balloccu, Schmidtová, Lango, and Dušek}]{balloccu2024leak}
Simone Balloccu, Patrícia Schmidtová, Mateusz Lango, and Ondřej Dušek. 2024.
\newblock \href {http://arxiv.org/abs/2402.03927} {Leak, cheat, repeat: Data contamination and evaluation malpractices in closed-source llms}.

\bibitem[{Braun(2023)}]{braun2023beg}
Daniel Braun. 2023.
\newblock I beg to differ: how disagreement is handled in the annotation of legal machine learning data sets.
\newblock \emph{Artificial Intelligence and Law}, pages 1--24.

\bibitem[{Braun and Matthes(2021)}]{braun-matthes-2021-nlp}
Daniel Braun and Florian Matthes. 2021.
\newblock \href {https://doi.org/10.18653/v1/2021.nlp4posimpact-1.10} {{NLP} for consumer protection: Battling illegal clauses in {G}erman terms and conditions in online shopping}.
\newblock In \emph{Proceedings of the 1st Workshop on NLP for Positive Impact}, pages 93--99, Online. Association for Computational Linguistics.

\bibitem[{Braun and Matthes(2022)}]{braun-matthes-2022-clause}
Daniel Braun and Florian Matthes. 2022.
\newblock \href {https://doi.org/10.18653/v1/2022.ecnlp-1.23} {Clause topic classification in {G}erman and {E}nglish standard form contracts}.
\newblock In \emph{Proceedings of the Fifth Workshop on e-Commerce and NLP (ECNLP 5)}, pages 199--209, Dublin, Ireland. Association for Computational Linguistics.

\bibitem[{Braun et~al.(2019)Braun, Scepankova, Holl, and Matthes}]{braun2019potential}
Daniel Braun, Elena Scepankova, Patrick Holl, and Florian Matthes. 2019.
\newblock The potential of customer-centered legaltech: Consumer protection in the digital era.
\newblock \emph{Datenschutz und Datensicherheit-DuD}, 43(12):760--766.

\bibitem[{Chalkidis et~al.(2019)Chalkidis, Androutsopoulos, and Aletras}]{chalkidis-etal-2019-neural}
Ilias Chalkidis, Ion Androutsopoulos, and Nikolaos Aletras. 2019.
\newblock \href {https://doi.org/10.18653/v1/P19-1424} {Neural legal judgment prediction in {E}nglish}.
\newblock In \emph{Proceedings of the 57th Annual Meeting of the Association for Computational Linguistics}, pages 4317--4323, Florence, Italy. Association for Computational Linguistics.

\bibitem[{Chalkidis et~al.(2017)Chalkidis, Androutsopoulos, and Michos}]{10.1145/3086512.3086515}
Ilias Chalkidis, Ion Androutsopoulos, and Achilleas Michos. 2017.
\newblock \href {https://doi.org/10.1145/3086512.3086515} {Extracting contract elements}.
\newblock In \emph{Proceedings of the 16th Edition of the International Conference on Articial Intelligence and Law}, ICAIL '17, page 19–28, New York, NY, USA. Association for Computing Machinery.

\bibitem[{Chalkidis et~al.(2020)Chalkidis, Fergadiotis, Malakasiotis, Aletras, and Androutsopoulos}]{chalkidis-etal-2020-legal}
Ilias Chalkidis, Manos Fergadiotis, Prodromos Malakasiotis, Nikolaos Aletras, and Ion Androutsopoulos. 2020.
\newblock \href {https://doi.org/10.18653/v1/2020.findings-emnlp.261} {{LEGAL}-{BERT}: The muppets straight out of law school}.
\newblock In \emph{Findings of the Association for Computational Linguistics: EMNLP 2020}, pages 2898--2904, Online. Association for Computational Linguistics.

\bibitem[{Chalkidis et~al.(2022)Chalkidis, Jana, Hartung, Bommarito, Androutsopoulos, Katz, and Aletras}]{chalkidis-etal-2022-lexglue}
Ilias Chalkidis, Abhik Jana, Dirk Hartung, Michael Bommarito, Ion Androutsopoulos, Daniel Katz, and Nikolaos Aletras. 2022.
\newblock \href {https://doi.org/10.18653/v1/2022.acl-long.297} {{L}ex{GLUE}: A benchmark dataset for legal language understanding in {E}nglish}.
\newblock In \emph{Proceedings of the 60th Annual Meeting of the Association for Computational Linguistics (Volume 1: Long Papers)}, pages 4310--4330, Dublin, Ireland. Association for Computational Linguistics.

\bibitem[{Chan et~al.(2020)Chan, Schweter, and M{\"o}ller}]{chan-etal-2020-germans}
Branden Chan, Stefan Schweter, and Timo M{\"o}ller. 2020.
\newblock \href {https://doi.org/10.18653/v1/2020.coling-main.598} {{G}erman{'}s next language model}.
\newblock In \emph{Proceedings of the 28th International Conference on Computational Linguistics}, pages 6788--6796, Barcelona, Spain (Online). International Committee on Computational Linguistics.

\bibitem[{Choi et~al.(2021)Choi, Hickman, Monahan, and Schwarcz}]{choi2021chatgpt}
Jonathan~H Choi, Kristin~E Hickman, Amy~B Monahan, and Daniel Schwarcz. 2021.
\newblock Chatgpt goes to law school.
\newblock \emph{J. Legal Educ.}, 71:387.

\bibitem[{Conneau et~al.(2019)Conneau, Khandelwal, Goyal, Chaudhary, Wenzek, Guzm{\'{a}}n, Grave, Ott, Zettlemoyer, and Stoyanov}]{DBLP:journals/corr/abs-1911-02116}
Alexis Conneau, Kartikay Khandelwal, Naman Goyal, Vishrav Chaudhary, Guillaume Wenzek, Francisco Guzm{\'{a}}n, Edouard Grave, Myle Ott, Luke Zettlemoyer, and Veselin Stoyanov. 2019.
\newblock \href {http://arxiv.org/abs/1911.02116} {Unsupervised cross-lingual representation learning at scale}.
\newblock \emph{CoRR}, abs/1911.02116.

\bibitem[{Drawzeski et~al.(2021)Drawzeski, Galassi, Jablonowska, Lagioia, Lippi, Micklitz, Sartor, Tagiuri, and Torroni}]{drawzeski-etal-2021-corpus}
Kasper Drawzeski, Andrea Galassi, Agnieszka Jablonowska, Francesca Lagioia, Marco Lippi, Hans~Wolfgang Micklitz, Giovanni Sartor, Giacomo Tagiuri, and Paolo Torroni. 2021.
\newblock \href {https://doi.org/10.18653/v1/2021.nllp-1.1} {A corpus for multilingual analysis of online terms of service}.
\newblock In \emph{Proceedings of the Natural Legal Language Processing Workshop 2021}, pages 1--8, Punta Cana, Dominican Republic. Association for Computational Linguistics.

\bibitem[{Gebru et~al.(2021)Gebru, Morgenstern, Vecchione, Vaughan, Wallach, au2, and Crawford}]{gebru2021datasheets}
Timnit Gebru, Jamie Morgenstern, Briana Vecchione, Jennifer~Wortman Vaughan, Hanna Wallach, Hal Daumé~III au2, and Kate Crawford. 2021.
\newblock \href {http://arxiv.org/abs/1803.09010} {Datasheets for datasets}.

\bibitem[{Hendrycks et~al.(2021)Hendrycks, Burns, Chen, and Ball}]{hendrycks2021cuad}
Dan Hendrycks, Collin Burns, Anya Chen, and Spencer Ball. 2021.
\newblock \href {https://openreview.net/forum?id=7l1Ygs3Bamw} {{CUAD}: An expert-annotated {NLP} dataset for legal contract review}.
\newblock In \emph{Thirty-fifth Conference on Neural Information Processing Systems Datasets and Benchmarks Track (Round 1)}.

\bibitem[{Katz et~al.(2023)Katz, Bommarito, Gao, and Arredondo}]{katz2023gpt}
Daniel~Martin Katz, Michael~James Bommarito, Shang Gao, and Pablo Arredondo. 2023.
\newblock Gpt-4 passes the bar exam.
\newblock \emph{Available at SSRN 4389233}.

\bibitem[{Leitner et~al.(2019)Leitner, Rehm, and Moreno-Schneider}]{leitner2019fine}
Elena Leitner, Georg Rehm, and Julian Moreno-Schneider. 2019.
\newblock {Fine-grained Named Entity Recognition in Legal Documents}.
\newblock In \emph{Semantic Systems. The Power of AI and Knowledge Graphs. Proceedings of the 15th International Conference (SEMANTiCS 2019)}, pages 272--287.

\bibitem[{Liu et~al.(2008)Liu, Wu, and Zhou}]{liu2008exploratory}
Xu-Ying Liu, Jianxin Wu, and Zhi-Hua Zhou. 2008.
\newblock Exploratory undersampling for class-imbalance learning.
\newblock \emph{IEEE Transactions on Systems, Man, and Cybernetics, Part B (Cybernetics)}, 39(2):539--550.

\bibitem[{Martin et~al.(2024)Martin, Whitehouse, Yiu, Catterson, and Perera}]{martin2024better}
Lauren Martin, Nick Whitehouse, Stephanie Yiu, Lizzie Catterson, and Rivindu Perera. 2024.
\newblock \href {http://arxiv.org/abs/2401.16212} {Better call gpt, comparing large language models against lawyers}.

\bibitem[{Minixhofer(2021)}]{Minixhofer_GerPT2_German_large_2020}
Benjamin Minixhofer. 2021.
\newblock \href {https://doi.org/10.5281/zenodo.5509984} {Gerpt2: German large and small versions of gpt2.}

\bibitem[{Ostendorff et~al.(2020)Ostendorff, Blume, and Ostendorff}]{10.1145/3383583.3398616}
Malte Ostendorff, Till Blume, and Saskia Ostendorff. 2020.
\newblock \href {https://doi.org/10.1145/3383583.3398616} {Towards an open platform for legal information}.
\newblock In \emph{Proceedings of the ACM/IEEE Joint Conference on Digital Libraries in 2020}, JCDL '20, page 385–388, New York, NY, USA. Association for Computing Machinery.

\bibitem[{Ruggeri et~al.(2022)Ruggeri, Lagioia, Lippi, and Torroni}]{ruggeri2022detecting}
Federico Ruggeri, Francesca Lagioia, Marco Lippi, and Paolo Torroni. 2022.
\newblock Detecting and explaining unfairness in consumer contracts through memory networks.
\newblock \emph{Artificial Intelligence and Law}, 30(1):59--92.

\bibitem[{Skadi{\c{n}}{\v{s}} et~al.(2014)Skadi{\c{n}}{\v{s}}, Tiedemann, Rozis, and Deksne}]{skadins-etal-2014-billions}
Raivis Skadi{\c{n}}{\v{s}}, J{\"o}rg Tiedemann, Roberts Rozis, and Daiga Deksne. 2014.
\newblock \href {http://www.lrec-conf.org/proceedings/lrec2014/pdf/846_Paper.pdf} {Billions of parallel words for free: Building and using the {EU} bookshop corpus}.
\newblock In \emph{Proceedings of the Ninth International Conference on Language Resources and Evaluation ({LREC}'14)}, pages 1850--1855, Reykjavik, Iceland. European Language Resources Association (ELRA).

\bibitem[{Torre et~al.(2020)Torre, Abualhaija, Sabetzadeh, Briand, Baetens, Goes, and Forastier}]{9218152}
Damiano Torre, Sallam Abualhaija, Mehrdad Sabetzadeh, Lionel Briand, Katrien Baetens, Peter Goes, and Sylvie Forastier. 2020.
\newblock \href {https://doi.org/10.1109/RE48521.2020.00025} {An ai-assisted approach for checking the completeness of privacy policies against gdpr}.
\newblock In \emph{2020 IEEE 28th International Requirements Engineering Conference (RE)}, pages 136--146.

\bibitem[{Wilson et~al.(2016)Wilson, Schaub, Dara, Liu, Cherivirala, Giovanni~Leon, Schaarup~Andersen, Zimmeck, Sathyendra, Russell, Norton, Hovy, Reidenberg, and Sadeh}]{wilson-etal-2016-creation}
Shomir Wilson, Florian Schaub, Aswarth~Abhilash Dara, Frederick Liu, Sushain Cherivirala, Pedro Giovanni~Leon, Mads Schaarup~Andersen, Sebastian Zimmeck, Kanthashree~Mysore Sathyendra, N.~Cameron Russell, Thomas~B. Norton, Eduard Hovy, Joel Reidenberg, and Norman Sadeh. 2016.
\newblock \href {https://doi.org/10.18653/v1/P16-1126} {The creation and analysis of a website privacy policy corpus}.
\newblock In \emph{Proceedings of the 54th Annual Meeting of the Association for Computational Linguistics (Volume 1: Long Papers)}, pages 1330--1340, Berlin, Germany. Association for Computational Linguistics.

\bibitem[{Younes and Mathiak(2022)}]{younes-mathiak-2022-handling}
Yousef Younes and Brigitte Mathiak. 2022.
\newblock \href {https://aclanthology.org/2022.icnlsp-1.9} {Handling class imbalance when detecting dataset mentions with pre-trained language models}.
\newblock In \emph{Proceedings of the 5th International Conference on Natural Language and Speech Processing (ICNLSP 2022)}, pages 79--88, Trento, Italy. Association for Computational Linguistics.

\end{thebibliography}

\clearpage
\appendix

\section{Taxonomy for Clause Topics}

See Table \ref{tab:topics}.

\label{sec:apptax}
\begin{table*}
\centering
\small
\begin{tabular}{|l l p{7cm}|}
\hline\textbf{Subtopic} & \textbf{Subtopic} & \textbf{Description}
\\\hline
\begin{filecontents*}{files/Taxonomy.csv}
Class,Subclass,Description,Keywords DE,Keywords EN,Ruletype
age,,Minimum age to order,,age,number
applicability,,Applicability of the T\&C,,,
applicableLaw,,Applicable law,,,keywordlist
arbitration,,Participation in arbitration,,,boolean
changes,,Changes to the contract,,,
codeOfConduct,,Code of conduct,Kodex,code of conduct,keywordlist
conclusionOfContract,,Conclusion of contract,,,
conclusionOfContract,binding,When the contract becomes binding,,,
conclusionOfContract,changeOfOrder,Changes and adujstments of orders,,,
conclusionOfContract,definition,Definition of terms used in the contract,,,
conclusionOfContract,restrictions,Restrictions to orders (e.g. the amount of ordered goods),,,
conclusionOfContract,steps,Steps towards contract conclusion,,,
conclusionOfContract,withdrawal,Withdrawal of the company from the contract,,,
delivery,,Delivery,,,
delivery,brokenPackaging,Handling of broken packaging ,,,
delivery,costs,Costs of delivery,,,
delivery,customs,Customs handling,,,
delivery,destination,Destinations to which goods are delivered,,,keywordlist
delivery,methods,Delivery methods,,,
delivery,partial,Partial delivery,,,
delivery,time,Delivery duration,,,number
description,,Product descriptions,,,
disposal,,Disposal regulations,,,
intellectualProperty,,Intellectual property,,,
language,,Contract language,Sprache, Vertragssprache,language, contract language,keywordlist
liability,,Liability,,,
party,,Contracting party,,,
payment,,Payment,,,
payment,fee,Payment fees,,,
payment,late,Late payment,,,
payment,loyalty,Loyalty schemes and reward programs,,,
payment,methods,Accepted payment methods,,,keywordlist
payment,restraint,Restraint of payment,,,
payment,vouchers,Vouchers,,,
personalData,,Personal data,,,
personalData,cookies,Cookie regulations,,,number
personalData,duration,Storage duration for personal data,,,number
personalData,information,Which information is processed/stored,,,
personalData,reason,Reason for storing / processing personal data,,,
personalData,update,Updates of personal data,,,
personalData,usage,Usage of personal data,,,
placeOfJurisdiction,,Place of jurisdiction,,,keywords
prices,,Prices,,,
prices,currency,Currency of prices,,,
prices,vat,VAT,,,
retentionOfTitle,,Retention of title,Eigentumsvorbehalt,,
severability,,Severability clause,salvatorisch,,
textStorage,,Storage of contract text,Speicherung Vertragstext,Storage contract text,boolean
warranty,,Warranty,,,
warranty,options,Options in case of warranty,,,keywordlist
warranty,period,Warranty period,,,number
withdrawal,,Withdrawal,,,
withdrawal,compensation,Compensation for product usage,,,
withdrawal,effects,Effects of withdrawal,,,
withdrawal,exclusion,Cases excluded from the right to withdraw,,,keywordlist
withdrawal,form,Allowed / disallowed to submit a withdrawal,,,keywordlist
withdrawal,model,Withdrawal model form,,,
withdrawal,period,Time period for withdrawal,,,number
withdrawal,shippingCosts,Shipping costs for withdrawal,,,
withdrawal,shippingMethod,Shipping method for withdrawal,,,keywordlist
\end{filecontents*}
\csvreader[before line=\\,before first line=]{files/Taxonomy.csv}{}{\csvcoli & \csvcolii & \csvcoliii}%
\\\hline
\end{tabular}
\caption{Taxonomy for Clause Topics (based on \citet{braun-matthes-2022-clause})}
\label{tab:topics}
\end{table*}

\section{Hyperparameters for Fine-Tuning}
\label{sec:appfine}

\subsection{BERT}
\begin{lstlisting}[caption={BERT parameters \texttt{agb-de} dataset}]
learning_rate = 2e-5,
per_device_train_batch_size = 2,
per_device_eval_batch_size= 2,
num_train_epochs = 5,
weight_decay= 0.01
\end{lstlisting}
\begin{lstlisting}[caption={BERT parameters \texttt{agb-de-under} dataset}]
learning_rate = 2e-5,
per_device_train_batch_size = 2,
per_device_eval_batch_size= 2,
num_train_epochs = 4,
weight_decay= 0.01
\end{lstlisting}
\subsection{RoBERTa}
\begin{lstlisting}[caption={RoBERTa parameters \texttt{agb-de} dataset}]
learning_rate = 2e-5,
per_device_train_batch_size = 8,
per_device_eval_batch_size= 8,
num_train_epochs = 3,
weight_decay= 0.01
\end{lstlisting}
\begin{lstlisting}[caption={RoBERTa parameters \texttt{agb-de-under} dataset}]
learning_rate = 2e-5,
per_device_train_batch_size = 2,
per_device_eval_batch_size= 2,
num_train_epochs = 4,
weight_decay= 0.01
\end{lstlisting}
\subsection{GPT2}
\begin{lstlisting}[caption={GPT2 parameters \texttt{agb-de} dataset}]
learning_rate = 2e-5,
per_device_train_batch_size = 2,
per_device_eval_batch_size= 2,
num_train_epochs = 6,
weight_decay= 0.01
\end{lstlisting}
\begin{lstlisting}[caption={GPT2 parameters \texttt{agb-de-under} dataset}]
learning_rate = 2e-5,
per_device_train_batch_size = 4,
per_device_eval_batch_size= 4,
num_train_epochs = 4,
weight_decay= 0.01
\end{lstlisting}

\section{GPT-3.5-Turbo Prompt}
\label{sec:appprompt}

\lstset{literate=%
    {Ö}{{\"O}}1
    {Ä}{{\"A}}1
    {Ü}{{\"U}}1
    {ß}{{\ss}}1
    {ü}{{\"u}}1
    {ä}{{\"a}}1
    {ö}{{\"o}}1
    {~}{{\textasciitilde}}1
}

\begin{lstlisting}[caption={GPT-3.5-Turbo Prompt and API call}]
model="gpt-3.5-turbo-0125",
response_format= { "type": "json_object" },
messages=[
      {"role": "system", "content": 'Stell dir vor, du bist ein Anwalt für Verbraucherschutz und berätst Verbraucher. Ist folgende Klausel in den AGB eines Online Shops potenziell ungültig wenn es sich beim Kunden um einen Verbraucher und beim Anbieter um eine Unternehmen handelt? Antworte mit true, wenn die Klausel potenziell ungültig ist und mit false wenn sie wahrscheinlich nicht ungültig ist. Erkläre deine Entscheidung. Antworte in folgendem JSON Format {"id": ' + str(id) + ', "ungültig": Boolean, "erklärung": String}'},
      {"role": "user", "content": text}
]
\end{lstlisting}

\section{Evaluation of undersampled models}
\label{sec:evalout}
\begin{table}[H]
\small
    \centering
    \begin{tabular}{|l|r|r|r|}
    \hline
         \textbf{Model} & \textbf{Precision} & \textbf{Recall} & \textbf{F1-score}  \\\hline
         \texttt{svm} & 0.20 & 0.32 & 0.25\\
         \texttt{bert} & 0.36 & 0.38 & 0.37\\
         \texttt{xlm} & 0.43 & 0.08 & 0.14\\
         \texttt{gerpt2} & 0.36& 0.43 & 0.40\\\hline
    \end{tabular}
    \caption{Evaluation results of the models trained on the undersampled training data on the full test data}
    \label{tab:evalout}
\end{table}

\section{GPT-3.5-Turbo Explanation Examples}
\subsection{Example 1}
\label{sec:apex1}
``Die Klausel ist potenziell ungültig, da sie den Verbraucher unangemessen benachteiligen könnte. Nach dem Gesetz muss der Verbraucher sein Widerrufsrecht klar und deutlich ausüben können, ohne dass zusätzliche Hürden oder Bedingungen auferlegt werden. Indem hier festgelegt wird, dass die Anmeldung einer Retoure unter Mein Konto als Widerruf gilt, könnte dies den Verbraucher in der Ausübung seines Widerrufsrechts einschränken. Es ist wichtig, dass der Verbraucher sein Widerrufsrecht ohne zusätzliche Verpflichtungen ausüben kann.''

\subsection{Example 2}
\label{sec:apex2}
``Gemäß § 357 BGB darf einem Verbraucher im Falle eines Widerrufs nicht die Kosten der Rücksendung auferlegt werden. Daher ist eine Klausel, die den Verbraucher verpflichtet, die Kosten für die Rücksendung zu tragen, potenziell ungültig.''

\newpage

\section{Datasheet}
\label{sec:datasheet}

\subsection{Motivation for Dataset Creation}

\textcolor{blue}{\textbf{Why was the dataset created?} (e.g., were there specific
tasks in mind, or a specific gap that needed to be filled?)}

The dataset was created to enable the training and evaluation of machine learning models that can detect potentially void clauses in consumer standard form contracts.

\textcolor{blue}{\textbf{What (other) tasks could the dataset be used for?} Are
there obvious tasks for which it should not be used?}

The dataset can also be used for clause topic classification.

\textcolor{blue}{\textbf{Has the dataset been used for any tasks already?} If so,
where are the results so others can compare (e.g., links to
published papers)?}

This paper is the first to use the dataset.

\textcolor{blue}{\textbf{Who funded the creation of the dataset?} If there is an
associated grant, provide the grant number.}

The data collection and annotation was supported by funds of the Federal Ministry of Justice and Consumer Protection (BMJV) based on a decision of the Parliament of the Federal Republic of Germany via the Federal Office for Agriculture and Food (BLE) under the innovation support
programme.

\subsection{Dataset Composition}

\textcolor{blue}{\textbf{What are the instances?} (that is, examples; e.g., documents, images, people, countries) Are there multiple types of instances? (e.g., movies, users, ratings; people, interactions between them; nodes, edges)}

Each instance consists of a clause from a consumer standard form contract and includes the text of the clause, the title (if any), the language of the clause, a unique ID, a unique ID identifying the contract the clause is from and three annotations: whether the clause was considered as potentially void by the annotators and a list of topics and subtopics.

\textcolor{blue}{\textbf{Are relationships between instances made explicit in
the data (e.g., social network links, user/movie ratings, etc.)?}}

Clause from the same contract are linked through the contract ID.

\textcolor{blue}{\textbf{How many instances of each type are there?}}

The dataset consists of 3764 clauses in total, 179 have been annotated as potentially void and and 3585 as likely valid.

\textcolor{blue}{\textbf{What data does each instance consist of?} “Raw” data
(e.g., unprocessed text or images)? Features/attributes? Is there a label/target associated with instances? If the
instances are related to people, are subpopulations identified
(e.g., by age, gender, etc.) and what is their distribution?}

Each instance consists of the clause text, the title of the clause (if any), the language of the clause, a unique ID, a unique ID identifying the contract the clause is from and three annotations: whether the clause was considered as potentially void by the annotators, a list of topics, and a list of subtopics.

\textcolor{blue}{\textbf{Is everything included or does the data rely on external
resources?} (e.g., websites, tweets, datasets) If external
resources, a) are there guarantees that they will exist, and
remain constant, over time; b) is there an official archival
version. Are there licenses, fees or rights associated with
any of the data?}

Everything is included in the dataset.

\textcolor{blue}{\textbf{Are there recommended data splits or evaluation measures?} (e.g., training, development, testing; accuracy/AUC)}

Splits for training and test are available together with the corpus. We suggest using metrics that work well on unbalanced data and highly discourage the use of accuracy as metric on this dataset.

\textcolor{blue}{\textbf{What experiments were initially run on this dataset?}
Have a summary of those results and, if available, provide
the link to a paper with more information here.}

The dataset was initially used to train classifiers that are able to detect potentially void clauses in consumer contracts.

\subsection{Data Collection Process}

\textcolor{blue}{\textbf{How was the data collected?} (e.g., hardware apparatus/sensor, manual human curation, software program, software interface/API; how were these constructs/measures/methods validated?)}

The data was manually collected by human annotators and copied into a structured Excel format.

\textcolor{blue}{\textbf{Who was involved in the data collection process?} (e.g.,
students, crowdworkers) How were they compensated? (e.g.,
how much were crowdworkers paid?)}

The data was collected by fully-qualified lawyers during their usual work-time. All participants worked for organizations that pay according to the collective labor agreement for public service workers in German states.

\textcolor{blue}{\textbf{Over what time-frame was the data collected?} Does the
collection time-frame match the creation time-frame?}

The data was collected between 2021 and 2022 and annotated between 2021 and 2023. The creation date of most of the items is unknown.

\textcolor{blue}{\textbf{How was the data associated with each instance acquired?} Was the data directly observable (e.g., raw text,
movie ratings), reported by subjects (e.g., survey responses),
or indirectly inferred/derived from other data (e.g., part of
speech tags; model-based guesses for age or language)? If
the latter two, were they validated/verified and if so how?}

The data was directly observable or was manually annotated by the annotators who are experts in the subject of the annotation.

\textcolor{blue}{\textbf{Does the dataset contain all possible instances?} Or is
it, for instance, a sample (not necessarily random) from a
larger set of instances?}

No, the dataset does not claim completeness in any sense.

\textcolor{blue}{\textbf{If the dataset is a sample, then what is the population?}
What was the sampling strategy (e.g., deterministic, probabilistic with specific sampling probabilities)? Is the sample representative of the larger set (e.g., geographic coverage)?
If not, why not (e.g., to cover a more diverse range of instances)? How does this affect possible uses?}

We believe that the dataset is somewhat representative for standard form consumer contracts in Germany. It is sampled from different industry (e.g. e-commerce and fitness). 

\textcolor{blue}{\textbf{Is there information missing from the dataset and why?}
(this does not include intentionally dropped instances; it
might include, e.g., redacted text, withheld documents) Is
this data missing because it was unavailable?}

The data has been anonymised, i.e. company names, phone numbers, addresses, tax ids, and similar information has been removed.

\subsection{Dataset Distribution}

\textcolor{blue}{\textbf{How is the dataset distributed?} (e.g., website, API, etc.;
does the data have a DOI; is it archived redundantly?)}

It is archived on GitHub (\url{https://github.com/DaBr01/AGB-DE}) and for easier access also available in the Hugging Face Hub (\url{https://huggingface.co/datasets/d4br4/agb-de}).

\textcolor{blue}{\textbf{When will the dataset be released/first distributed?} (Is
there a canonical paper/reference for this dataset?)}

June 2024.

\textcolor{blue}{\textbf{What license (if any) is it distributed under?} Are there
any copyrights on the data?}

The annotations are licensed under CC-BY-SA 4.0.

\textcolor{blue}{\textbf{Are there any fees or access/export restrictions?}}

No.

\subsection{Dataset Maintenance}
\textcolor{blue}{\textbf{Who is supporting/hosting/maintaining the dataset?}
How does one contact the owner/curator/manager of the
dataset (e.g. email address, or other contact info)?}

Daniel Braun, \url{d.braun@utwente.nl}

\textcolor{blue}{\textbf{Will the dataset be updated?} How often and by whom?
How will updates/revisions be documented and communicated (e.g., mailing list, GitHub)? Is there an erratum?}

There are no plans to update the dataset unless important mistakes become clear.

\textcolor{blue}{\textbf{If the dataset becomes obsolete how will this be communicated?}}

On the GitHub page.

\textcolor{blue}{\textbf{Is there a repository to link to any/all papers/systems
that use this dataset?}}

Yes.

\textcolor{blue}{\textbf{If others want to extend/augment/build on this dataset,
is there a mechanism for them to do so?} If so, is there
a process for tracking/assessing the quality of those contributions. What is the process for communicating/distributing
these contributions to users?}

We would suggest to create a fork on GitHub.

\subsection{Legal \& Ethical Considerations}

\textcolor{blue}{\textbf{If the dataset relates to people (e.g., their attributes) or
was generated by people, were they informed about the
data collection?} (e.g., datasets that collect writing, photos,
interactions, transactions, etc.)}

There is no information about individuals in the data or was recorded during the annotation of the data.

\textcolor{blue}{\textbf{If it relates to other ethically protected subjects, have
appropriate obligations been met?} (e.g., medical data
might include information collected from animals)}

N.a.

\textcolor{blue}{\textbf{If it relates to people, were there any ethical review applications/reviews/approvals?} (e.g. Institutional Review
Board applications)}

N.a.

\textcolor{blue}{\textbf{If it relates to people, were they told what the dataset
would be used for and did they consent? What community norms exist for data collected from human communications?} If consent was obtained, how? Were the people
provided with any mechanism to revoke their consent in the
future or for certain uses?}

N.a.

\textcolor{blue}{\textbf{If it relates to people, could this dataset expose people
to harm or legal action?} (e.g., financial social or otherwise)
What was done to mitigate or reduce the potential for harm?}

N.a.

\textcolor{blue}{\textbf{If it relates to people, does it unfairly advantage or disadvantage a particular social group?} In what ways? How
was this mitigated?}

N.a.

\textcolor{blue}{\textbf{If it relates to people, were they provided with privacy
guarantees?} If so, what guarantees and how are these
ensured?}

N.a.

\textcolor{blue}{\textbf{Does the dataset comply with the EU General Data Protection Regulation (GDPR)?} Does it comply with any other
standards, such as the US Equal Employment Opportunity
Act?}

Yes, since only publicly available information was collected, the dataset complies with the GDPR and similar regulations.

\textcolor{blue}{\textbf{Does the dataset contain information that might be considered sensitive or confidential?} (e.g., personally identifying information)}

No.

\textcolor{blue}{\textbf{Does the dataset contain information that might be considered inappropriate or offensive?}}

No.

\end{document}